\documentclass[conference]{IEEEtran}
\IEEEoverridecommandlockouts
\usepackage{cite}
\usepackage{amsmath,amssymb,amsfonts}
\usepackage{algorithmic}
\usepackage{graphicx}
\usepackage{textcomp}
\usepackage{xcolor}

\usepackage[caption=false,font=normalsize,labelfont=sf,textfont=sf]{subfig}
\usepackage{stfloats}
\usepackage{tikz}
\usepackage{algorithm}
\usepackage{tabularx, booktabs}
\def\BibTeX{{\rm B\kern-.05em{\sc i\kern-.025em b}\kern-.08em
    T\kern-.1667em\lower.7ex\hbox{E}\kern-.125emX}}

\newtheorem{thm}{Theorem}[section]
\newtheorem{lem}{Lemma}[section]

\newtheorem{rem}{Remark}[section]

\begin{document}

\title{A Vectorization Method Induced By Maximal Margin Classification For Persistent Diagrams\\
\thanks{* Corresponding author}
}

\author{
\IEEEauthorblockN{An Wu \textsuperscript{*}}
\IEEEauthorblockA{\textit{College of Mathematics and Computer Science} \\
\textit{Zhejiang A$\&$F University}\\
Hangzhou, China \\
20220075@zafu.edu.cn}
\and
\IEEEauthorblockN{Yu Pan}
\IEEEauthorblockA{\textit{School of Mathematics and Statistics} \\
\textit{Beijing Institute of Technology}\\
Beijing, China \\
p4nyu@foxmail.com}
\and
\IEEEauthorblockN{Fuqi Zhou}
\IEEEauthorblockA{\textit{School of Electronics and Information} \\
\textit{Hangzhou Dianzi University}\\
Hangzhou, China \\
23040444@hdu.edu.cn}
\and
\IEEEauthorblockN{Jinghui Yan}
\IEEEauthorblockA{\textit{Institute of Automation} \\
\textit{Chinese Academy of Sciences}\\
Beijing, China \\
jinghui.yan@nlpr.ia.ac.cn}
\and
\IEEEauthorblockN{Chuanlu Liu}
\IEEEauthorblockA{\textit{School of Computer Science $\&$ Technology} \\
\textit{Beijing Institute of Technology}\\
Beijing, China \\
chuanluliu@163.com}
}

\maketitle

\begin{abstract}
Persistent homology is an effective method for extracting topological information, represented as persistent diagrams, of spatial structure data. Hence it is well-suited for the study of protein structures. Attempts to incorporate Persistent homology in machine learning methods of protein function prediction have resulted in several techniques for vectorizing persistent diagrams. However, current vectorization methods are excessively artificial and cannot ensure the effective utilization of information or the rationality of the methods. To address this problem, we propose a more geometrical vectorization method of persistent diagrams based on maximal margin classification for Banach space, and additionaly propose a framework that utilizes topological data analysis to identify proteins with specific functions. We evaluated our vectorization method using a binary classification task on proteins and compared it with the statistical methods that exhibit the best performance among thirteen commonly used vectorization methods. The experimental results indicate that our approach surpasses the statistical methods in both robustness and precision.
\end{abstract}

\begin{IEEEkeywords}
Maximal margin classification, Topological data analysis, Persistent diagram, Vectorization, Protein function
\end{IEEEkeywords}

\section{Introduction}
Topological data analysis (TDA) is a computational technique which has been developed to study the topological structure of data over the past 20 years\cite{C09,C20}. The theoretical foundation of TDA is the algebraic topology, and combined with computer technology TDA has been applied to various practical problems, such as image processing, material structure analysis and quantum computing to name a few as in \cite{PC14,CLW23,HNH16} and \cite{LGZ16}.

\begin{figure*}[ht]
    \centering
    \includegraphics[width=0.7\textwidth]{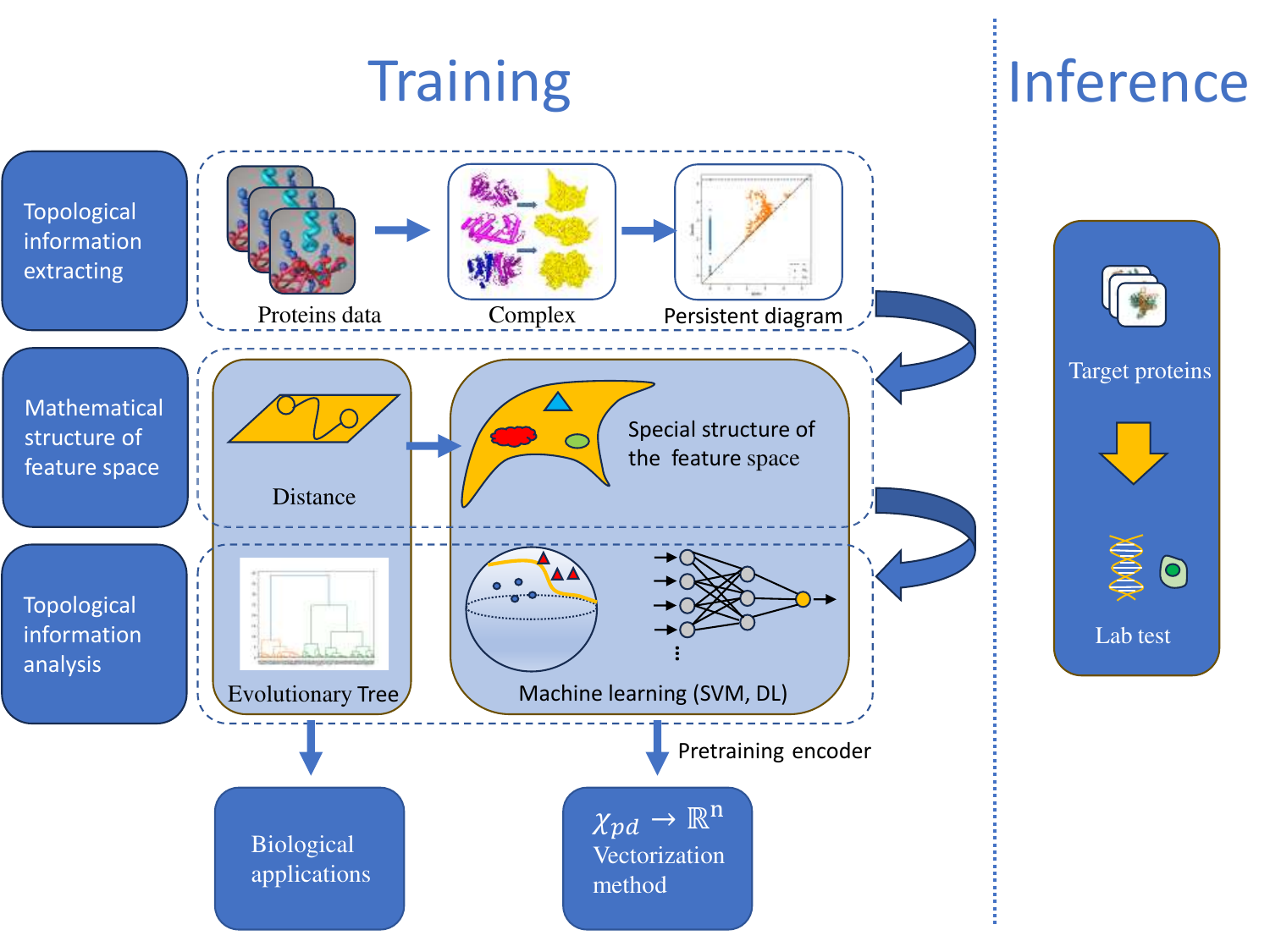}
    \caption{Framework of protein functional prediction by TDA}
    \label{process}
\end{figure*}

The primary methods in Topological Data Analysis are persistent homology and Mapper, and we focus on the persistent homology approach in this paper. Based on the persistent homology there are many interesting works. The research team led by Rongling Wu implemented GLMY homology theory proposed by Shing-Tung Yau to analyze and interpret the topological change of health state from symbiosis to dysbiosis, and they successfully described the topological difference and commonality between Crohn's disease and ulcerative colitis\cite{WLD23}. Perea introduced how to combine dynamical systems and topological data analysis to gain insights from time-varying data\cite{P19}. For more applications of persistent homology, one can refer to the review paper of Chazal and Michel\cite{CM21}. Previous works have primarily focused on classification problems as an application of persistent homology. This is because the number of bars in the barcode (Betti numbers) offers a direct measure for classification, or because researchers can define similar functions (such as bottleneck distance and p-Wasserstein distance) based on the persistent diagrams. However, if we want to combine persistent diagrams with more advanced artificial intelligence technologies, there is a critical problem that most machine learning algorithms assume that the input data is from $\mathbb{R}^n$ or, more generally, some Hilbert space, but the space of persistent diagrams is not a vector space even if they are from the similar samples. Hence we want to find a  reasonable embedding of persistent diagrams which can be a pre-encoder for any other topological artificial intelligence frameworks.

Many researchers have tried to solve this problem which is called vectorization. Barnes, Polanco and Perea provided a comparative study of six such methods and they applied and compared popular machine learning methods on five data sets\cite{BPP21}. Chung and Lawson developed a framework of vectorization and showed several well-known summaries fall in this framework\cite{CL22}. In \cite{DHZ20} and \cite{DLZ24}, stable vector representations were developed and these methods were tested on several data sets to show their effectiveness and efficiency. Ali et al. \cite{AAJ23} summarized thirteen vectorization methods, and they examined these methods on three well-known classification tasks. They discovered that the best-performing method is a simple vectorization by a few statistics. Besides these methods, there are numerous approaches of vectorization as in\cite{CNO20,K18} and \cite{ZLJ20}, and more information can be found in Section 2.3 of \cite{ZKN24}. If we further consider the reasons behind Ali and his colleagues’ results, a plausible explanation is that overly artificial vectorization methods may lead to the loss of key information, whereas statistical methods are relatively more simple and straightforward. However, statistical methods are also hard to interpret. For example, it is difficult to answer why someone need to take the 10th, 25th, 75th and 90th percentiles of the births and deaths rather than others. Hence we approve a more explainable method which is not included in the methods mentioned above, and our method achieve a better result than the statistical method \cite{AAJ23} which achieves the best performance on several datasets.

As previously stated, one can define a distance function on persistent diagrams to construct a metric space. Consequently, a direct approach is performing classification tasks within this metric space. But a primary challenge mentioned in \cite{AAJ23} of this method is that 
\textit{"The space of persistent diagrams is infinite-dimensional and highly nonlinear",}
which poses significant computational and theoretical complexities. However, in practice, since the size of data is always finite, the discrete points in the persistent diagram are bounded. Therefore, we can always construct the  compactification of the space of persistent diagrams generated by the sample set. Finally, by the classical functional analysis theory \cite{HBS05}, we have
\begin{lem}
A compact metric space can always be embedded into a Banach space.
\end{lem}
Then we can solve the challenge above on a Banach space.

In the current paper, we focus on the space $(\mathcal{X}_{pd}, d)$ generated by finite persistent diagrams endowed with a fixed metric. Then inspired by the theory of maximal margin classification for Banach spaces \cite{HBS05}, we embed $(\mathcal{X}_{pd}, d)$ into the Banach space $\mathcal{C}_b(\mathcal{X}_{pd})$ by Kuratowski's embedding, then we consider the maximal margin classification for space $\mathcal{C}_b(\mathcal{X}_{pd})$. As a result, we transform the maximum margin problem into a classical quadratic programming problem as Eq. (\ref{qpp}). Moreover, in the derivation process, we find the following result:
\begin{thm}
Given a set of persistent diagram $\mathcal{X}_{pd}$ endowed with metric $d$, we define the vectorization $\mathcal{X}_{pd}\rightarrow \mathbb{R}^n$ by $x\mapsto (d(x,x_1),\cdots, d(x,x_n))$, where $x_i\in \mathcal{X}_{pd}$ and $n$ is the cardinal number of $\mathcal{X}_{pd}$. In addition, this simple vectorization method corresponds to the quadratic programming problem Eq. (\ref{qpp}). 
\end{thm} 

Next, we apply the maximal margin classification for persistent diagrams on the data of Cas-associated proteins, which are closely related to gene editing technology as in\cite{CRC13} and \cite{JCF12}, and transposases. We choose a simple distance defined by Eq. (\ref{dis}) which requires less computation than the bottleneck distance, and we compare our method with statistical methods which are the best-performing methods in the thirteen vectorization methods \cite{AAJ23} mentioned before. As a consequent, our method achieves superior performance both in robust and accuracy. 

Finally, we present a framework that uses the TDA processing methodology introduced in this paper to predict proteins with specific functions, and the total framework is exhibited in Fig. \ref{process}. We will discuss more details in Section III.

The remainder of our paper is organized as follows. In section II, we revisit the maximal margin classification for Banach spaces and apply it to the space of persistent diagrams. Finally, the problem is reduced to a solvable quadratic programming problem. In section III, we show our experiments on the data of Cas-associated proteins and  transposases. We analysis our experiment results and propose a framework of predicting proteins' function. In section IV, we conclusion our work and do some discussions on the current and future researches.

\section{Maximal Margin Classification Method}
In this section, we will introduce the main idea of TDA first. Then we will review the principle of the maximal margin classification for Banach spaces, and one can check the details in the publication of Hein, Bousquet and Sch\"{o}lkopf\cite{HBS05}. In particular, we will regard the space of persistent diagrams as a Banach space, which allows us to analyze the space of persistent diagrams by maximal margin classification. 

\subsection{The topological data analysis}
\begin{figure}[t]
\centering
\begin{tikzpicture}[scale=0.3]
\coordinate (A1) at (-2.4, 4.3);
\coordinate (A2) at (-2.6, 2.9);
\coordinate (A3) at (-1, 3.5);
\coordinate (A4) at (1.3, -0.08);
\coordinate (A5) at (2.14, -1.14);

\foreach \p in {A1,A2,A3,A4,A5} {
    \filldraw (\p) circle (2pt);
}

\begin{scope}[xshift=5cm]
\coordinate (B1) at (-2.4+5, 4.3);
\coordinate (B2) at (-2.6+5, 2.9);
\coordinate (B3) at (-1+5, 3.5);
\coordinate (B4) at (1.3+5, -0.08);
\coordinate (B5) at (2.14+5, -1.14);

\foreach \p in {B1,B2,B3,B4,B5} {
    \filldraw (\p) circle (2pt);
    \filldraw[fill=none, draw=red] (\p) circle (15pt);
}
\end{scope}

\begin{scope}[xshift=10cm]
\coordinate (C1) at (-2.4+10, 4.3);
\coordinate (C2) at (-2.6+10, 2.9);
\coordinate (C3) at (-1+10, 3.5);
\coordinate (C4) at (1.3+10, -0.08);
\coordinate (C5) at (2.14+10, -1.14);

\draw (C1)--(C2);
\draw (C2)--(C3);
\draw (C1)--(C3);
\draw (C4)--(C5);

\foreach \p in {C1,C2,C3,C4,C5} {
    \filldraw (\p) circle (2pt);
    \filldraw[fill=none, draw=red] (\p) circle (24pt);
}
\end{scope}
\end{tikzpicture}
\caption{An Intuitive Explanation of Persistent Homology}\label{persistent_homology}
\end{figure}
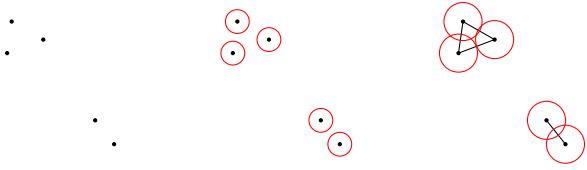

The main idea of TDA is to find some topological features to describe the difference of data. As an example, assume we have a deflated balloon. When we inflate it, it remains a balloon, because this process is a continuous transformation. However, if we blow in enough gas to cause an explosion, it ceases to be a balloon and becomes a ”burst balloon”. Hence the topological feature that distinguishes burst balloons from intact balloons is the presence or absence of holes in the balloons. By popularizing this idea, researchers have developed a major technique in TDA called persistent homology\cite{ELZ02,C09}. The word "persistent" means that there is a continuous variation of a parameter which is call the filtration. If we can construct a series of complexes suitable with the continuous variation, then the homological information of the complexes illustrate the topological structure of data. For example, Fig. \ref{persistent_homology} shows the process of increasing the radius of points, and it is easy to find that the topological structure of top left three points is different to that of bottom right two points, as the former would create a hole.

\begin{figure}[b]
    \centering
    \includegraphics[width=0.5\textwidth]{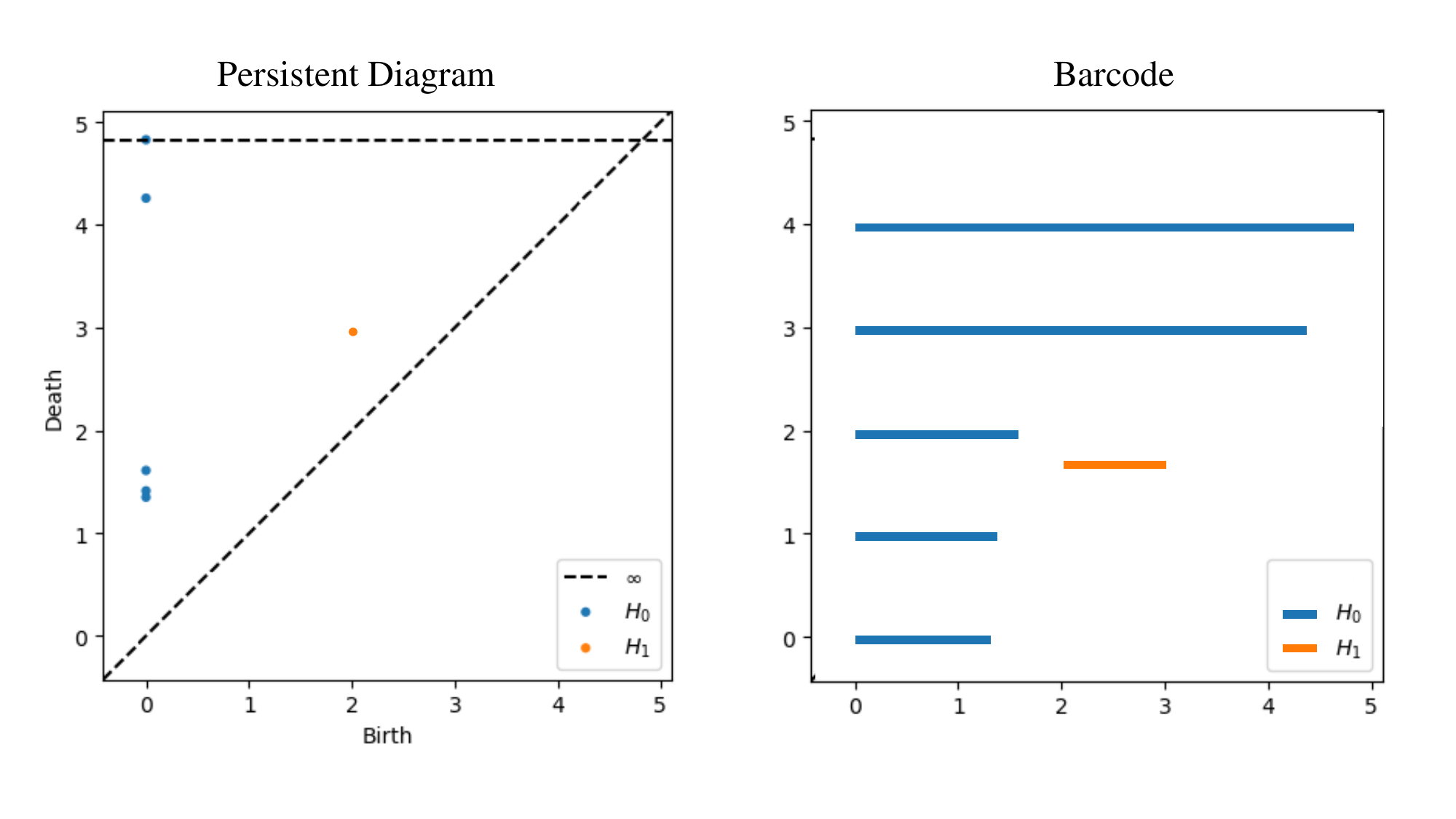}
    \caption{Visualizations of topological features}
    \label{pd_bar1}
\end{figure}

Based on this idea, persistent homology is good at dealing with classification tasks, and the topological features are usually taken as the connect components, holes, and voids which are denoted by $H_i, i=1,2,3$ respectively. Typically, there are two visualization methods for this information, known as the persistent diagram and the barcode, which are essentially equivalent to each other. In Fig. \ref{pd_bar1}, we show both visualizations of topological information extracted from Fig. \ref{persistent_homology}. 
\begin{rem}
The construction of complexes in persistent homology is of importance. In this section, we use the Čech complex to illustrate the geometric explanation. In practice, the Vietoris-Rips complex is often employed, as it is beneficial for computational purposes.
\end{rem}

However, the number of points contained within different persistent diagrams may indeed differ. Hence, we need a vectorization method to transfer persistent diagrams into a vector space, and we will illustrate our vectorization method next.

\subsection{The maximal margin classification}
Considering the space of persistent diagrams $\mathcal{X}_{pd}$, we can define a distance function $d(\cdot, \cdot)$ on $\mathcal{X}_{pd}$ to make it a metric space $(\mathcal{X}_{pd},d)$. In practical problems, the number of generated persistent diagrams is always finite, then we can assume that $d(X, Y)$ is bounded for any sample $X, Y\in\mathcal{X}_{pd}$, and we use the same symbol $\mathcal{X}_{pd}$ to denote the subspace generated by sample set. Hence we obtain a compact metric space $(\mathcal{X}_{pd}, d)$, which can be embedded into a Banach space isometrically by the observation of Kuratowski as in\cite{HBS05, JJ19}.

Let $\mathcal{C}_b(\mathcal{X}_{pd})$ be the Banach space of continuous and bounded functions on $\mathcal{X}_{pd}$ endowed with infinite norm $\|.\|_{\infty}$. Take an arbitrary point $x_0 \in \mathcal{X}_{pd}$ and define a map from $\mathcal{X}_{pd}$ to real-valued functions on $\mathcal{X}_{pd}$  
\begin{equation}
\phi : x\mapsto \phi_x:=d(x, \cdot)-d(x_0,\cdot).  
\end{equation}
Then consider the space $D :=\overline{\mathop{\text{span}}\{\phi_x:x\in \mathcal{X}_{pd}\}}$, in which the closure is taken in $(\mathcal{C}_b(\mathcal{X}_{pd}), \|.\|_{\infty})$, we have the following lemma: 
\begin{lem}\label{kuratowski}
  $\phi$ is a total isometric embedding from $(\mathcal{X}_{pd},d)$ into the Bananch space $(D, \|.\|_{\infty})\subset (\mathcal{C}_b(\mathcal{X}_{pd}), \|.\|_{\infty})$.
\end{lem}

Through Lemma \ref{kuratowski}, we can transform the data processing problem on space $(\mathcal{X}_{pd}, d)$ to that on space $(D, \|.\|_{\infty})$.  In the current paper, we focus on the maximal margin classification of persistent diagrams. Recall that, the margin formulation on a Banach space is
\begin{align}\label{optW'}
& \mathop{\text{min}}\limits_{W'\in D', b\in \mathbb{R}} \|W'\|\nonumber\\
& s.t. \  y_j(\langle W', \phi_{x_j}\rangle_{D',D}+b)\geq 1, \forall j=1,\cdots, n,
\end{align}
where $D'$ is the dual space of $D$, $y_j$ is the label corresponding to $\phi_{x_j}$ and the product $\langle W', \phi_{x_j}\rangle_{D', D}$ is naturally defined by $W'(\phi_{x_j})$.

Since $\mathcal{X}_{pd}$ is compact, $\mathcal{C}_b(\mathcal{X}_{pd})$ is isometrically isomorphic to the Banach space $\mathcal{M}(\mathcal{X}_{pd})$ of finite signed Borel measures on $\mathcal{X}_{pd}$ with the measure norm. However, we did not construct the parametrization of $\mathcal{M}(\mathcal{X}_{pd})$, hence we need to choose a reasonable approximation on dual space $\mathcal{C}_b(\mathcal{X}_{pd})'$ instead.

Define a space $E:=\mathop{\text{span}}\{\delta_z: z\in \mathcal{X}_{pd}\}$, where $\delta_z(f):=f(z)$ for any $f\in \mathcal{C}_b(\mathcal{X}_{pd})$. There is the following lemma:
\begin{lem}\label{E}
  The space $E$ is weak*-dense in $\mathcal{C}_b(\mathcal{X}_{pd})'$ endowed with the norm $$\|\sum\limits_{i=1}^n\beta_i\delta_{z_i}\|_{\mathcal{C}_b(\mathcal{X}_{pd})'}=\sum\limits_{i=1}^n|\beta_i|.$$
\end{lem}
Because of the Lemma \ref{E}, we can use $e=\sum\limits_{i=1}^n\beta_i\delta_{z_i}\in E$ to approximate any elements of $\mathcal{C}_b(\mathcal{X}_{pd})'$. Substitute $e$ into $W'$ in Eq. (\ref{optW'}), the optimization problem can be rewritten as 
\begin{align}\label{opte}
& \mathop{\text{min}}\limits_{n\in \mathbb{N}, z_i\in \mathcal{X}_{pd}, b} \sum\limits_{i=1}^n |\beta_i|\nonumber\\
& s.t. \  y_j(\sum\limits_{i=1}^n\beta_i(d(x_j, z_i)-d(x_0, z_i))+b)\geq 1.
\end{align}
In consideration of computational convenience, practically it is necessary to restrict $E$ to the training set, which means $n$ is the size of training set. Moreover, it can be noted that the term $-\sum\limits_{i=1}^n\beta_id(x_0, z_i)+b$ is independent of $j$, so that we can reparameterize it as $c$. In addition, the vector $\beta=(\beta_1, \cdots, \beta_n)$ is optimized in a $n$ dimensional vector space, where dimension $n$ is always finite. Hence the 1-norm in the optimization function is equivalent to 2-norm, which is more advantageous in solving our problem. In light of these observation, we convert the optimization problem as Eq. (\ref{opte}) into solving the following optimization problem:   
\begin{align}\label{opte2}
& \mathop{\text{min}}\limits_{\beta_i, c} \sum\limits_{i=1}^n \|\beta_i\|^2\nonumber\\
& s.t. \  y_j(\sum\limits_{i=1}^n\beta_id(x_j, z_i)+c)\geq 1, \forall j=1,\cdots, n.
\end{align}
Observe the form of the optimization problem above, one can find it exhibits a form identical to that of the support vector machine defined in Euclidean space. 

Now we will subsequently explain this phenomenon from the perspective of data science. It is obvious that the primary difference lies in the replacement of term $\sum\limits_{i=1}^n\beta_ix_j$ in SVM with term $\sum\limits_{i=1}^n\beta_id(x_j, z_i)$ in the conditions. Because in SVM the data $x_j$ is always a vector, but now $x_j$ is just a point in the sample space which may no longer be a vector space, therefore we can not do linear combination of samples here directly. However, for each pair of sample points $(x_j, z_i)$ we can define the distance (or similarity) function between them to make the sample space be a metric space. Then for each sample we can generate a fixed-length vector by computing distances from this sample to all points in training set. Essentially, we transform the metric space into a linear space through the distance function, and this transformation method is theoretically guaranteed to be non-arbitrary in mathematics.

Same as SVM, we can also introduce slack variables to obtain a new version with a soft margin as follows:
\begin{align}\label{soft}
\mathop{\text{min}}\limits_{\beta_i, \xi_i, c} & \sum\limits_{i=1}^n \|\beta_i\|^2+a\sum\limits_{i=1}^n\xi_i\nonumber\\
s.t. \  & y_j(\sum\limits_{i=1}^n\beta_id(x_j, z_i)+c)\geq 1-\xi_j, \nonumber\\
& \xi_j\geq 0, \forall j=1,\cdots, n,
\end{align}
where $\xi_j$ is the slack variable which is defined by hinge loss 
$$\xi_j=\mathop{\text{max}}(0, 1-y_j(\sum\limits_{i=1}^n\beta_id(x_j, z_i)+c)).$$
In the current paper, we focus on the soft margin version of maximal margin classification, and solving Eq. (\ref{soft}) is a typical quadratic programming problem. Here, we render it in matrix notation for ease of subsequent resolution.

Define the coefficient vector $\beta:=(\beta_1, \cdots, \beta_n)^T$, the slack vector $\xi:=(\xi_1,\cdots,\xi_n)^T$, label vector $y:=(y_1, \cdots, y_n)^T$ and denote the distance matrix of training set by $d$. Then the optimization function is represented by
\begin{equation}
\mathop{\text{min}}\limits_{\beta,\xi, c} \frac{1}{2}\beta^T2\mathop{\text{I}}\nolimits_n\beta+a \mathbf{1}_n^T\xi,
\end{equation}
where $\mathop{\text{I}}_n$ is identity matrix with $n$ order and $\mathbf{1}_n^T$ is a $n$-dimensional vector with elements are all equal to $1$. Furthermore, combine $\beta$, $\xi$, and $c$ together, we obtain
\begin{equation}
\mathop{\text{min}}\limits_{\alpha} \frac{1}{2}\alpha^TQ\alpha+b^T\alpha,
\end{equation}
where $\alpha, Q, b$ are defined as follows:
\begin{align}
  & \alpha := \left(\begin{array}{c}
  \beta \\
  \xi \\
  c
  \end{array}\right), \ b :=\left(\begin{array}{c}
  0\\
  a\mathbf{1}_n\\
  0
  \end{array}\right), \nonumber\\
  & Q := \left(\begin{array}{ccc}
                2\mathop{\text{I}}_n & 0 & 0 \\
                0 & 0 & 0 \\
                0 & 0 & 0 
              \end{array}
  \right).\nonumber
\end{align}
If we denote the Hadamard product of matrices by the symbol $*$.
Then similarly, the matrix representation of constraint conditions is
\begin{equation}
G\alpha \geq h,
\end{equation}
where the matrix $G$ and the vector $h$ are defined by
\begin{equation}
G:=\left(\begin{array}{ccc}
           y*d & \mathop{\text{I}}_n & \mathbf{1}_n \\
           0 & \mathop{\text{I}}_n & 0 
         \end{array}\right), \  
h:=\left(\begin{array}{c}
\mathbf{1}_n\\
0
\end{array}\right).\nonumber
\end{equation}

In a conclusion, we transform the problem of maximal margin classification on the space of persistent diagrams into solving the quadratic programming problem
\begin{align}\label{qpp}
& \mathop{\text{min}}\limits_{\alpha} \frac{1}{2}\alpha^TQ\alpha+b^T\alpha,\nonumber\\
& s.t. \    G\alpha \geq h,
\end{align}
which can be solved directly.

\section{Experiments and Analysis}
In this section, we will illustrate our experiments of maximal margin classification of persistent diagrams. We will apply our Banach space based method, which will be referred as BS method in the following context, to the classification of proteins. Then we will compare the experiment results of BS method with two widely used methods which are based on dimension reduction techniques. Finally, I will analyze the significance of BS method and its applications in the exploration of novel proteins.

\subsection{Experiment Setup}
Our experimental setup employs a server equipped with an Intel® Xeon® CPU E7-4870 operating at a frequency of 2.40 GHz, and we use the Python library \textit{ripser} \cite{B21, TSB18} to compute persistent homology. For solving the quadratic programming problem, we use the quadratic programming solver \textit{CVXOPT}.

\subsection{Dataset}
The original dataset we deal with consists of $161$ protein PDB files of Cas-associated proteins and $197$ files of transposases, and our goal is to classify these proteins correctly. Each PDB file includes the coordinates of atoms in the proteins, and it is easy to extract these coordinate information from PDB files. After extracting the atomic coordinates, the traditional approach of TDA is computing the persistent homology of the atomic point clouds directly. However, this methodology encounters computational challenges, particularly in the calculation of the second persistent homology group $(H_2)$, which is often exceedingly difficult and necessitates substantial computational expenditures. The main reason is that compared to general chemical molecules and structural materials, proteins are typically composed of tens of thousands of atoms. But when calculating homology information of $H_2$, it is necessary to traverse all combinations of four atomic interactions. Therefore, in this paper, we modify the persistent homology method at the beginning.

\subsection{Preprocessing}
We extract the topological information of protein atomic point clouds according to the following approach. Firstly, we derive the coordinates of 358 proteins from the input files. Subsequently, we construct the residue connection matrix of each protein based on the relative positions of the C$\alpha$ atoms. For more details, we consider the C$\alpha$ atoms within a protein as nodes, and an edge is created between two C$\alpha$ atoms if the inter-atomic distance is below this established threshold of 5 angstroms (Å). Following the construction of  residue connection matrices, we can use graph embedding methods to generate point clouds which include less points than the original atomic point clouds. There are many graph embedding methods, and here we choose \textit{Node2vec} \cite{GL16} method to generate new point clouds. After the embedding of residue connection matrices, we calculate the persistent homology of point clouds including $H_0, H_1$ and $H_2$. With the topological features captured, the final step is preprocessing of these topological information. We set a cutoff of 0.01 to filter out $H_1$ and $H_2$ information whose persistent times are shorter than this value. We summarize the topological feature extraction algorithm as Algorithm 1.

\begin{algorithm}[H]
\caption{Mortified Persistent Homology}\label{alg:alg1}
\begin{algorithmic}
\STATE 
\STATE {\textsc{Input:}} Protein PDB files $\{f_1, \cdots, f_{358}\}$.
\STATE {\textsc{Process:}} 
\STATE \    \textbf{for} $j=1,\cdots,358$ \textbf{do}
\STATE \     \  extract point cloud $PC_j$ of $f_j$;
\STATE \     \  construct the residue connection matrix $G_j$ of $PC_j$;
\STATE \     \  generate graph embedding $\overline{PC_j}$ of $G_j$;
\STATE \     \  compute persistent homology information $PD_j$ of $\overline{PC_j}$;
\STATE \     \  filter out noise information.
\STATE \    \textbf{end for}
\STATE {\textsc{Output:}} Persistent diagrams $\{PD_1, \cdots, PD_{358}\}$.
\end{algorithmic}
\label{alg1}
\end{algorithm}

By using the residue connection matrix and graph embedding, we reduce the number of points in the point clouds. Hence, it is foreseeable that the aforementioned approach to extracting topological features will have a substantial time-saving advantage. In addition, because we only use the inter-atomic distances of the C$\alpha$ atoms, some information is inevitably omitted, which is acceptable in our experiment.

The results of preprocess is that the computation time required for calculating $H_1, H_2$, and $H_3$ using the modified persistent homology approach is comparable to the time taken for calculating $H_1$ and $H_2$ using the original persistent homology method, with both processes approximately lasting around $10$ days. Roughly speaking, this implies that the new persistent homology algorithm omits certain two-dimensional topological information, and the computational resources saved are redirected towards the calculation of three-dimensional topological information. Although not the focus of this paper, it is worth mentioning that this strategy has a distinct advantage in that it introduces three-dimensional information that was previously unavailable, which will finally provide an additional weight parameter for learning in other machine learning processes.

\subsection{Baselines and Main Method}
After the modified persistent homology, we will get persistent diagrams with three components $H_1, H_2$ and $H_3$ which are essentially point sets. Next, we will introduce three widely used topological features analysis methods which are based on different statistics as baselines for our BS method as follows:
\begin{itemize}
\item Statistical Method I\cite{CLW23, WLD23}: We count the cardinality of each point set for persistent diagram $PD_i$, which is equivalent to the persistent Betti number, directly to construct a $3$-dimensional feature vector $v_i$. And the first element is normalized by multiplying with a factor $0.01$. 
\item Statistical Method II: 
    We compute the statistical information (the maximum, the minimum, the variance, the mean value and the median) for $5$ indicators, corresponding to the birth or death times of $H_1, H_2$ and $H_3$, to generate the feature vector of each persistent diagram. Then we extract a $25$-dimensional feature vector for each persistent diagram. 
\item Statistical Method III: We apply the feature vectors proposed in the Definition 3.1 of the survey paper \cite{AAJ23}. This statistical vectorization method incorporates more statistical information than the two methods mentioned above, and it is the best-performing method of the thirteen alternative methods in three well-known classification tasks.
\end{itemize}

In comparison with the statistical methods, the BS method needs a distance function $d$ to make the space of persistent diagrams $\mathcal{X}_{pd}$ be a metric space. In this paper, we define the distance function as follows:
\begin{equation}\label{dis}
d(X,Y):=\sum\limits_{i=1}^3 w_id_i(X, Y), \    \forall X,Y\in \mathcal{X}_{pd},
\end{equation}
where $d_i(X,Y)$ 
is the maximum Euclidean distance between $H_i$-components 
of persistent diagrams $X$ and $Y$, and $W=(w_1,w_2,w_3)$ is weight vector which is fixed as $(1/3, 1/3, 1/3)$ in our experiment.  

In order to compare the effects of different vectorization methods, we apply all methods on the dataset by soft-SVM to obtain classification results.

\subsection{Results and Analysis}
We perform five-fold cross-validation experiments and continuously increase the proportion of training data for all methods mentioned before. The experimental results are presented in the following Table I.

\begin{table}[tbp]\label{mth}
    \setlength{\tabcolsep}{2pt}
    \centering
    \caption{Accuracy of cross-validation experiments}
    \begin{tabular}{lccc} 
        \toprule
        \textbf{Methods} & \textbf{Thirty percent} & \textbf{Fifty percent} & \textbf{Eighty percent} \\
        \midrule
        \textbf{BS Method (our)} & 83.51\% & 84.47\% & \textbf{86.94\%}\\
        Statistical Method I & \textbf{83.90\%} & \textbf{85.47\%} & 84.44\%\\
        Statistical Method II & 81.12\% & 81.68\% & 85.28\%\\
        Statistical Method III & 81.35\% & 83.13\% & 85.56\%\\
        \bottomrule
    \end{tabular}
\end{table}

The accuracies of the BS method and Statistical methods are noted to increase with the expansion of the training set size. While the Statistical method I has relatively small change around 84\%-85\%, which means this method exhibits robustness. As a comparison, the Statistical method II and III are more sensitive to the size of training date, and these methods have lower accuracies in the case with small training set. It is worth mentioning that Statistical methods require manual feature selection, and this example demonstrates that different feature selection approaches have a significant impact on the accuracy. Therefore, it needs a prior understanding of the original data which will impact the feature selection. 

However, our BS method uses the information of all features and has a fixed processing procedure, which reduces errors that may arise from poor feature selection. Moreover, the BS method also has robustness, and when the training date size reaches eighty percent, BS method becomes the best method in accuracy. More importantly, the BS method is more generalizable. Because for any persistent diagrams with more topological information ($H_4, H_5,\cdots$), we can also define a distance function like Eq. (\ref{dis}) and we can apply the neural network to determine the weight vector $W$ in practice. 

\subsection{Case Study}
In the end, we will illustrate the classification of proteins in the view of biology and propose a framework of finding new proteins with certain functions by topological data analysis. 
\begin{figure}[tbp]
    \centering
    \includegraphics[width=0.2\textwidth]{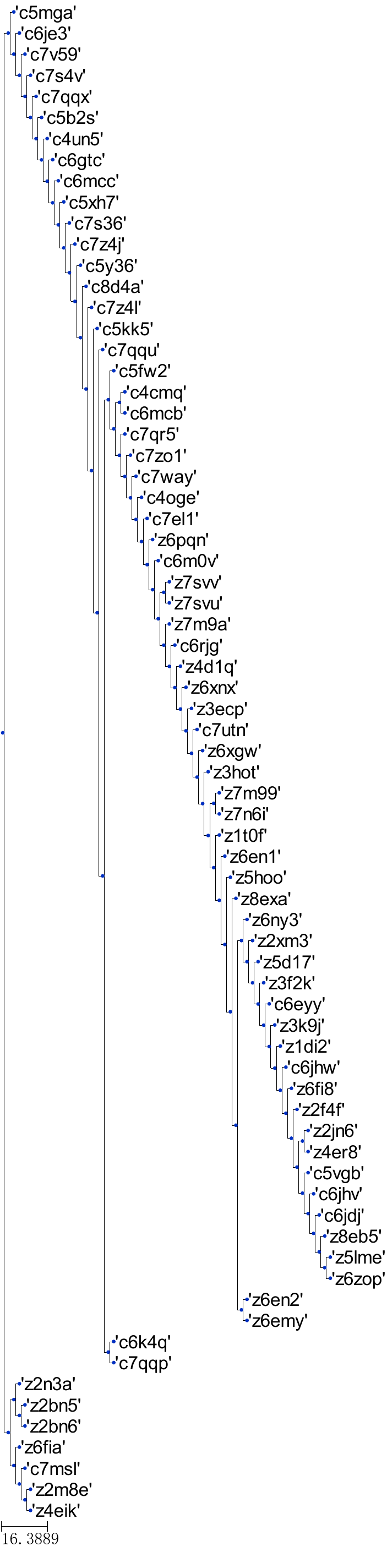}
    \caption{Evolution tree of 72 proteins}
    \label{tree}
\end{figure}

For giving a biological explanation, we take 80\% of the data to be the training set randomly and use the remaining data to test as well as to build an evolutionary tree. In Fig. \ref{tree}, we show the evolutionary tree and in experiment result we find that 7 proteins have been misclassified. These proteins are 'z7m9a', 'c5vgb', 'c6eyy', 'c6jhv', 'c6jhw', 'c7msl' and 'c7tun', in which we use the characters “c” and “z” to denote Cas-associated proteins and transposases respectively.

It is obvious that, Fig. \ref{tree} can be primarily divided into two components: the upper component mainly consists of Cas-associated proteins, while the lower component mainly consists of transposases. And we find that classification errors can be divided into two parts: 
\begin{itemize}
\item Type I: The first involves the occurrence of another class of proteins within a particular category (such as 'c5vgb', 'c6eyy', 'c6jhv', 'c6hjw' and 'c7msl').
\item Type II: The second involves the presence of errors in the crossover region in which both classes of proteins appear alternatively    (such as 'z7m9a' and 'c7utn'). 
\end{itemize}

However, if we focus on the instances of misclassification, we find that the reason for Type I error is that the sample classification can be further refined, whereas the reason for Type II error is that the sample classification is overly granular. In details, Cas-associated proteins can be further subdivided into Cas proteins and anti-Cas proteins. Particularly, we find that proteins 'c5vgb', 'c6eyy', 'c6jhv', 'c6jhw' and 'c7msl' are anti-Cas proteins as in\cite{HDM17, HRA18, KLY19} and \cite{HPG22} which are distinct from the majority of other Cas-associated proteins, hence they are misclassified. On the other hand, proteins 'z7m9a' and 'c7utn' are Cas proteins as well as transposases as in \cite{PTM21} and \cite{SHK22}, and the fact suggests that this dual nature has resulted in their misclassification. Therefore,  our classification methodology aligns with the established principles and demonstrates superior performance.

\subsection{Framework for Protein Function Prediction}
Moreover, Type II error approves a new method for discovering novel proteins with specific functionalities. For example, if a transposase in the crossover region is misclassified, we can suspect reasonably that it exhibits structural and functional features similar to those of Cas proteins. Finally, we summarize this framework for exploring protein functions by their spatial structures in Fig. \ref{process}:
\begin{itemize}
\item In the first step, we input the PDB files $\{f_j\}$ of proteins with a known specific function and $\{g_i\}$ of validation proteins;
\item In the second step, we calculate the topological information for every PDB file from its spatial structure;
\item In the third step, we conduct the binary classification task described in Section II to differentiate between the sets $\{f_j\}$ and $\{g_i\}$ using our vectorization method of persistent diagrams;
\item Finally, we confirm those proteins with the second type of error and identify the subset $\{g_k\}$ of proteins that exhibit this function.
\end{itemize} 

\section{Conclusion and Discussion}
\subsection{Conclusion}
We propose a novel vectorization method, which not only attains the best performance in processing topological information compared with conventional methods but also exhibits robustness against the size of the training dataset, to analysis persistent diagrams. We derive the process for handling persistent diagrams using the maximal margin classification method of a Banach space, and discover that once the metric space of persistent diagrams is embedded by Kuratowski's embedding then the final result is equivalent to an encoder that calculates the distances between a given persistent diagram and all other diagrams. Hence our result proposes a more reasonable feature selection which use more topological information extracted than statistical methods that are the best method between thirteen alternative
methods\cite{AAJ23}. 

Another result with more profound implications for biology is that we propose a framework 
of using spatial topological information of proteins to identify their functions. Because we find that in our binary classification experiment, the second type of error typically arises from proteins that belong to both classes. Therefore, to verify whether certain proteins possess a specific function, we need only focus on whether they would result in a Type II error when classified alongside proteins which are known to exhibit this function.

As a byproduct of our processing, we have optimized the persistent homology algorithm for protein analysis, therefore its computational efficiency is enhanced and it is permitted to calculate the second Betti number.

\subsection{Discussion}
In future work, we plan to investigate that whether the Banach space based method can be improved by following modifications:
\begin{itemize}
  \item First, we intend to examine the performance of our method across various distances;
  \item Second, we plan to combine our method with neural network methods;
  \item Finally, we aim to propose additional vectorization methods for persistent diagrams, informed by geometric perspectives
\end{itemize}

In the derivation process of the BS method, we used information from the distance function. 
Compared to traditional methods of comparing persistence diagrams, such as the bottleneck distance, 
the theoretical computational complexity of our method is lower 
because it does not require enumerating all possible matchings. 
However, we may get a better performance distance by training the metric matrix with the neural network.

Our method can be integrated with the neural network in more aspects, because our method maps proteins into a vector space of relative topological information, 
and this vector representation 
offers greater interpretability. In our subsequent work, we will combine the vectorization approach presented in this paper with neural networks to address further problems related to the spatial structure of proteins. 

Finally, since the vectorization method proposed in this paper originates from a geometric embedding, we aim to consider more vectorization methods inspired by other geometric structures such as Lie groups and homogeneous spaces. 
Particularly, we can choose different embedding methods than Kuratowski's embedding to construct different reasonable encoders for persistent diagrams. 

\section{Acknowledgements}
The authors  extend many thanks to Professor Chao Qian and Hanning Wang who provide academical support and encouragement.

\vspace{12pt}

\end{document}